# UltraLightSqueezeNet: A Deep Learning Architecture for Malaria Classification with up to 54x fewer trainable parameters for resource constrained devices

Suresh Babu Nettur[1*], Shanthi Karpurapu[1*], Unnati Nettur[2], Likhit Sagar Gajja[3], Sravanthy Myneni[1], Akhil Dusi[4] AND Lalithya Posham[5]

[1] Independent Researcher, Virginia Beach, VA 23456, USA
[3] Department of Computer Science, Virginia Tech, Blacksburg, VA 24061 USA
[3] Department of Computer Science, BML Munjal University, Haryana 122413 INDIA
[4] Department of Information Systems, University of Indiana Tech, Indiana USA
[5] Nanjing Medical University, Nanjing City, Jiyangsu, CHINA

Corresponding authors: Shanthi Karpurapu (shanthi.karpurapu@gmail.com), Suresh Babu Nettur (nettursuresh@gmail.com)

[*] Shanthi Karpurapu and Suresh Babu Nettur are co-first authors

**ABSTRACT** Lightweight deep learning approaches for malaria detection have gained attention for their potential to enhance diagnostics in resource constrained environments. For our study, we selected SqueezeNet1.1 as it is one of the most popular lightweight architectures. SqueezeNet1.1 is a later version of SqueezeNet1.0 and is 2.4 times more computationally efficient than the original model. We proposed and implemented three ultra-lightweight architecture variants to SqueezeNet1.1 architecture, namely Variant 1 (one fire module), Variant 2 (two fire modules), and Variant 3 (four fire modules), which are even more compact than SqueezeNetV1.1 (eight fire modules). These models were implemented to evaluate the best performing variant that achieves superior computational efficiency without sacrificing accuracy in malaria blood cell classification. The models were trained and evaluated using the NIH Malaria dataset. We assessed each model's performance based on metrics including accuracy, recall, precision, F1-score, and Area Under the Curve (AUC). The results show that the SqueezeNet1.1 model achieves the highest performance across all metrics, with a classification accuracy of 97.12%. Variant 3 (four fire modules) offers a competitive alternative, delivering almost identical results (accuracy 96.55%) with a 6x reduction in computational overhead compared to SqueezeNet1.1. Variant 2 and Variant 1 perform slightly lower than Variant 3, with Variant 2 (two fire modules) reducing computational overhead by 28x, and Variant 1 (one fire module) achieving a 54x reduction in trainable parameters compared to SqueezeNet1.1. These findings demonstrate that our SqueezeNet1.1 architecture variants provide a flexible approach to malaria detection, enabling the selection of a variant that balances resource constraints and performance.

## I. INTRODUCTION

Malaria, caused by the protozoan parasite *Plasmodium falciparum,* is a severe disease transmitted through the bites of infected mosquitoes. Symptoms include fever, headaches, nausea, and, in severe cases, seizures, jaundice, and coma, which can ultimately lead to death. In 2022, the World Health Organization reported approximately 249 million malaria cases and 608,000 fatalities worldwide. The African region accounts for approximately 94% of malaria-related deaths globally. Vulnerable populations, including pregnant women, children, infants under five, refugees, migrants, internally displaced people, and Indigenous Peoples [1], are disproportionately affected by the disease. This data highlights the critical necessity for improved strategies in malaria prevention and treatment.

In Ethiopia, malaria is a major health concern, affecting more than 68% of the population of over 100 million. The country experiences significant malaria epidemics every five to eight years, with annual focal outbreaks. Over 75% of the landscape below 2,000 meters in elevation is prone to malaria infections [2], leading to approximately 2.9 million cases reported annually [3]. This results in increased morbidity and mortality, especially during epidemics. Early detection and prompt treatment are crucial in reducing death rates associated with the disease. It is recommended that treatment be initiated within 24 hours of fever onset, particularly for vulnerable populations such as young children. Early intervention is fundamental in preventing severe complications and reducing mortality rates.

Traditional methods for identifying malaria-infected red blood cells, such as manual microscopic examination of Wright-Giemsa-stained slides, are prone to inaccuracies due to human error. While polymerase chain reaction (PCR) diagnostics are sensitive, they require PCR machines and reagents to conduct the assays. Other detection methods, such as traditional cell culture and pathogen isolation, can be tedious, time-consuming, costly, and often lack sensitivity [4].

In recent years, computer-aided diagnosis utilizing deep learning techniques has demonstrated promising advancements in medical imaging analysis, particularly in malaria classification. The integration of machine learning algorithms in malaria classification systems can significantly enhance automation, improve diagnostic accuracy, reduce treatment delays, and alleviate the workload on healthcare professionals. Computer vision techniques have achieved remarkable success across various medical image analysis tasks, including disease detection and diagnosis [5]. These approaches are mainly employed as computer-aided diagnostic (CAD) systems, providing rapid support and improving diagnostic accuracy [6]. Previous studies have concentrated on classifying and detecting malaria in cells using conventional methods, including Convolutional Neural Network (CNN) architectures such as LeNet [7], VGG [8], AlexNet [9], Inception [10], ResNet [11], and EfficientNet [12], alongside non-CNN methods like Support Vector Machine (SVM) [13] and XG-Boost [14] [15]. While these strategies have demonstrated effective performance in detecting and diagnosing malaria, there has been limited exploration into leveraging SqueezeNet mechanisms to enhance malaria disease detection.

The primary objective of our research is to drive advancements in the application of deep learning and LLMs within the fields of software engineering [16][17] and biomedical sciences [18][19].The objective of our current study is to develop lightweight deep learning models, as malaria remains a major global health challenge. To achieve this, we aimed to introduce and develop lightweight variants of SqueezeNet [20] to further reduce computational resource needs without compromising efficiency. We chose SqueezeNet as it is one of the best lightweight, high-performing CNN architectures known for requiring fewer computational resources. Our study utilizes the Kaggle Malaria dataset(sourced from NIH), a large, comprehensive, and well-annotated source of malaria cell images, making it an ideal foundation for training deep learning models tailored to this task.

## II. RELATED WORKS

Conventional machine learning (CML) methods have been widely utilized for automated malaria diagnosis over the years. However, since 2017, there has been a significant shift toward deep learning (DL) methods, owing to their advanced capabilities and improved effectiveness in malaria diagnosis compared to CML for Malaria Classification.

### A. DEEP LEARNING APPROACHES FOR MALARIA CLASSIFICATION

Previous studies have explored deep learning methods for detecting malaria in blood cells, yielding promising results. Dong et al. (2017) [21] investigated automated malaria diagnosis using deep learning, leveraging a dataset of red blood cell images labeled by pathologists. They evaluated LeNet, AlexNet, and GoogLeNet, achieving classification accuracies over 95%, surpassing the 92% accuracy of the support vector machine (SVM). The study highlighted the ability of deep learning methods to automatically extract features, reducing reliance on human expertise [21]. Bibin et al. (2017) [22] utilized a deep belief network (DBN) to classify 4,100 blood smear images as parasite or non-parasite, achieving an F-score of 89.66%, sensitivity of 97.60% and specificity of 95.92%. This first application of DBNs for malaria detection outperformed state-of-the-art methods using color and texture features with a pre-trained and fine-tuned DBN architecture. In a later study, Rajaraman et al. (2018) [23] evaluated pre-trained CNNs for malaria parasite detection in blood smear images, showing promising results for feature extraction and classification. Vijayalakshmi et al. (2019) [24] developed a deep learning model combining VGG19 and Support Vector Machine (SVM) for identifying falciparum malaria parasites. Using transfer learning, the model achieved 93.1% classification accuracy, outperforming state-of-the-art CNN models in accuracy, sensitivity, specificity, precision, and F-score. This approach highlights the potential of transfer learning in malaria diagnosis [24]. Cinar et al. (2020) [25] developed a CNN-based system to classify malaria images as healthy or parasitized, achieving 97.83% accuracy with DenseNet201 using Gaussian-filtered data.

Preprocessing with medium and Gaussian filters improved performance, and the Matlab implementation works independently of image size [25]. Abubakar et al. (2021) [26] developed a machine learning-based system for malaria parasite detection in blood-smear images using six feature extraction models (VGG16, VGG19, ResNet50, ResNet101, DenseNet121, DenseNet201) and classifiers like Decision Tree, SVM, Naïve Bayes, and K-NN. The system achieved over 94% accuracy with reduced complexity compared to previous methods, demonstrating its effectiveness for malaria detection [26]. Arshad et al. (2022) [27] developed a deep learning-based method for automatic malaria parasite detection using a two-stage approach: classifying cells as healthy or infected, then identifying life-cycle stages (ring, trophozoite, schizont, gametocyte). They applied ResNet50V2 and DenseNet201 and achieved 95.86% accuracy with ResNet50V2 [27]. Madhu et al. (2022) [28] developed an automated model for one-shot detection and classification of malaria thin blood smears using the Deep Siamese Capsule Network (D-SCN). Their approach, featuring a capsule network with an imperative routing mechanism and Lorentz similarity metric, achieved a detection accuracy of 97.24% and classification accuracy of 98.89%. This work is the first to apply D-SCN for malaria diagnosis with state-of-the-art performance [28]. Bhuiyan et al. (2023) [29] developed an ensemble learning-based deep learning model for malaria parasite detection in red blood cell images. Using VGG16, VGG19, and DenseNet201 models, they applied adaptive weighted average ensemble methods and max voting. Data augmentation was employed to address overfitting. The model outperformed others, achieving 97.92% accuracy, demonstrating its potential for accurate and automatic malaria diagnosis [29].

## B. LIGHTWEIGHT DEEP LEARNING APPROACHES FOR MALARIA CLASSIFICATION

Lightweight deep learning models have become essential in malaria classification due to their ability to operate effectively on devices with limited computational resources, such as mobile phones and embedded systems. Yang et al. [30] developed a smartphone app using deep learning for detecting malaria parasites in thick smear images, combining IGMS for screening and CNN for classification. Their method, the first to apply deep learning on smartphones for thick smears, is evaluated at the patient level [30]. Shivaramakrishnan et al. (2017) [31] developed a customized deep learning model for malaria cell classification, achieving 98.61% accuracy with lower complexity and computation time. Their model outperformed state-of-the-art methods, including pre-trained models, offering an effective tool for large-scale automated malaria diagnosis [31]. Magotra et al. [32] proposed a customized CNN architecture for malaria classification, which consists of six convolutional layers with filter sizes ranging from 16 to 128, followed by the use of the "Relu" activation function and six max-pooling layers.

The model consists of 381,729 parameters, with 381,313 (0.364 MB 8 bitwise) trainable parameters and 416 non-trainable parameters, balancing computational efficiency and classification performance. Chaudhary et al. [33] proposed a lightweight deep learning model for malaria parasite-type detection and life cycle stage classification, featuring fewer than 0.4 million parameters, making it over 20 times lighter than DenseNet. They tested on four different malaria datasets, and their model outperformed existing methods and was specifically designed for deployment in mobile applications, particularly in resource-limited settings. Al Muazzaz et al. [34] proposed a lightweight CNN model for early-stage malaria parasite detection from microscopic blood images, achieving 99.56% accuracy, 99.97% precision, 98.36% recall, 99% F1-score, and 99.2% AUC score. The model's efficiency was demonstrated with a prediction time of 2 milliseconds per image. Grad-CAM was used to highlight key image regions, improving interpretability. The model's lightweight nature makes it highly suitable for practical deployment in resource-limited settings, enhancing healthcare accessibility and efficiency. Zedda et al. (2022) [35] proposed a real-time malaria parasite detection and classification system using YOLOv5 and DarkNet-53 networks. Their study compared different CNN architectures for four-class classification of Plasmodium falciparum life stages. The results showed that YOLOv5 achieved 95.2% accuracy, and DarkNet-53 achieved 96.02%, outperforming state-of-the-art methods. This approach enables improvements in recognizing malaria parasite species and life stages, even in mobile environments [35]. Goni et al. [36] proposed a customized lightweight CNN with 0.17M parameters for malaria detection from RBC images, achieving 99.45% accuracy, 99.75% precision, 99.17% recall, and 99.46% F1-score.

The model, evaluated using SHAP for explainability, outperformed transfer learning and SOTA models in efficiency and performance, making it suitable for fast, reliable malaria detection. Khan et al. [37] proposed a fractional-order optimizer-based lightweight CNN for malaria diagnosis, addressing computational inefficiency and interpretability issues in existing models. The model achieved 95% accuracy on the NIH dataset [38] and demonstrated robustness with 92% and 90.4% accuracy on the MP-IDB and M5 test sets, respectively. Salam et al. [39] utilized a lightweight 17-layer SqueezeNet model with 119,154 trainable parameters for automatic malaria detection, leveraging an open-source dataset of 26,161 blood smear images. The model, optimized for embedded systems with a size of just 1.72 MB, achieved exceptional performance metrics on the original NIH dataset: 96.37% accuracy, 95.67% precision, 97.21% recall, and 96.44% F1 score.

Results improved to 99.71% across all metrics on a modified dataset. Tested on the Jetson Nano B01 edge device, it demonstrated a rapid prediction time of 0.24 seconds per image. This efficient and accurate model is highly suitable for malaria detection in resource-constrained settings. Shahadat et al. [40] proposed a parameter-efficient deep learning model for malaria detection using microscopic blood smear images. The architecture integrates a Squeeze-and-Excitation block with a spatial 1D CNN layer, achieving a testing accuracy of 99.52% on a modified dataset and 96.78% on the original dataset. With 2.2M trainable parameters and 12.6M FLOPS, the model delivers state-of-the-art performance, making it suitable for mobile and resource-constrained applications. Maqsood et al. [41] proposed a customized CNN model for malaria detection from thin blood smear images, using bilateral filtering for noise removal and image augmentation to improve generalization.

The model, with five convolutional layers, five max-pooling layers, and two fully connected layers, achieved 96.82% accuracy. It was computationally efficient, with an inferencing time of 5 seconds for a dataset of 8,272 images and a training time of 25 minutes for 25 epochs. Eze et al. [42] proposed a deep learning model for malaria detection in resource-constrained areas, achieving up to 99% recall with quantized versions of Basic Convolutional Neural Network (B-CNN) and MobileNetV2. The MobileNetV2 model, which was deployed in their mobile application, had a size of 2MB (8-bit quantized) and an inference time of 33–95 microseconds on mobile phones. This model was selected for its efficient memory usage and fast inference, making it ideal for deployment in energy-limited environments. Elangovan et al. [43] developed a computationally efficient 18-layer CNN for malaria parasite detection, focusing on reducing the number of learnable parameters and computational complexity. The approach outperformed pre-trained networks in terms of classification accuracy while significantly reducing computational resources, with simulation results showing an overall accuracy of 97.8% and an F1-score of 97.84%.

Our literature study shows a notable lack of exploration into the use of light weight CNNs like SqueezeNet for detecting malaria in blood smears. As a result, many existing approaches exhibit high computational costs. SqueezeNet is a highly efficient CNN yet compact architecture and is ideal for resource-constrained environments, such as mobile and embedded systems. Our work bridges this gap by introducing novel, lightweight variants of SqueezeNet designed to deliver high performance while even further minimizing computational resource requirements, making it an even more promising solution for real-world malaria detection.

## III. METHODOLOGY

Our proposed models for malaria detection adopt a well-structured methodology, as shown in Figure 1, carefully followed to ensure reliability and precision at every stage. The process begins with the acquisition and preprocessing of malaria cell images, where various image enhancement techniques are applied to improve data quality and relevance for analysis. Next, we implement the novel SqueezeNet1.1 architecture variants we introduced alongside the SqueezeNet1.1 architecture. These variants were developed to further reduce the complexity of the SqueezeNet1.1 architecture, aiming to lower computational resource requirements while maintaining high performance. The original SqueezeNet1.1 architecture was modified with 1, 2, and 4 fire modules and was systematically explored to optimize the balance between accuracy and efficiency. The selected variant is then trained on the preprocessed image dataset. Following training, the model undergoes rigorous evaluation to assess its effectiveness in distinguishing between malaria-infected and healthy cells. Finally, we compare the performance of our novel SqueezeNet1.1 architecture variants to identify the most suitable one that ensures better performance and resource efficient malaria detection.

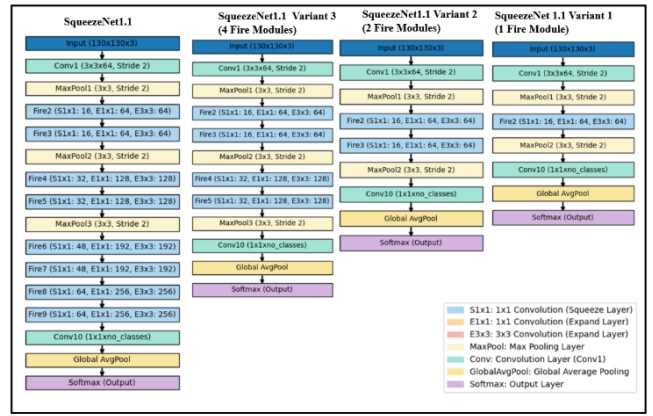

**FIGURE 1.** SqueezeNet1.1 and its architecture variants

*1) Dataset*

Our research leverages the malaria dataset [38] from Kaggle (sourced from NIH), a well-established and thoroughly annotated collection of microscopic images tailored for malaria-related research. The dataset includes 27,558 images, divided equally between 13,779 images of uninfected cells and 13,779 images of parasitized red blood cells. We allocated 20% (5,512 images) for validation and 80% (22,046 images) for training. Example images from the dataset are shown in Figure 2.

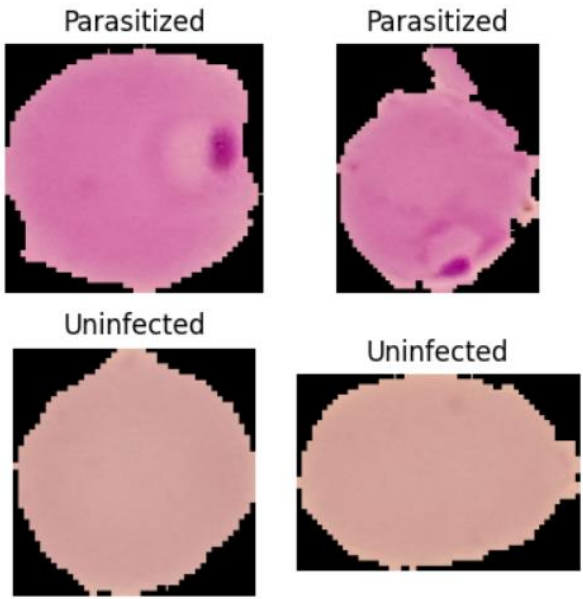

FIGURE 2. Sample Malaria images

*2) Image Processing*

In this study, we implemented image processing techniques to enhance the detection of malaria-infected cells in microscopic images. The approach included resizing images to a target size of (130, 130), converting them from BGR to RGB color space, and normalizing pixel values to a range of [0, 1] for consistent input to the model. These preprocessing steps standardize the images, ensuring consistency and improving the model's ability to learn effectively. Additionally, we applied image augmentation to increase the diversity of the training dataset and improve the model's generalization. The augmentation process included rotation (±10 degrees), zooming (up to 10%), width and height shifts (up to 10% of the image size), and flipping images both horizontally and vertically. By introducing these augmentations, we effectively simulated variability in the dataset, making the model more robust to real-world scenarios.

*3) SqueezeNet*

Smaller CNN models, like SqueezeNet, offer significant advantages, such as reduced computational overhead and faster training, making them well-suited for deployment on resource-limited devices. Their scalability across various hardware environments and ability to simplify cloud-to-device deployment enables efficient model downloads for autonomous systems. Additionally, they are ideal for memory-constrained hardware like Field-Programmable Gate Arrays (FPGAs), which excel in parallel processing within resource-limited contexts. Given these advantages, we selected SqueezeNet as the core architecture for our solution.

SqueezeNet, introduced by Iandola et al. in 2016 [20], is a CNN architecture particularly well-suited for tasks demanding efficient performance, such as image classification and real-time inference on mobile and edge devices [44] [45] [46] [47] [48] [49]. It achieves recognition precision comparable to AlexNet [50] while reducing the number of parameters to approximately 1/50th of its size.

The architecture introduces a unique building block called the fire module, as shown in Figure 3, which serves as the foundation for constructing lightweight and effective CNNs. The fire module consists of a squeeze layer and an expand layer, each designed to optimize parameter efficiency and feature representation. Each module includes a rectified linear unit (ReLU) activation function to enhance non-linearity and facilitate deeper network architectures. The squeeze layer applies 1x1 convolutions to reduce the number of input channels, while the expand layer employs a mix of 1x1 and 3x3 convolutions to enhance the network's representational capacity. The use of 1x1 filters in the expanding layer is crucial for minimizing computational cost, whereas the inclusion of 3x3 filters enables the network to capture spatially local features effectively. The architecture of SqueezeNet is built upon three fundamental design strategies, as shown in Figure 4, aimed at reducing parameter count while maintaining competitive accuracy.

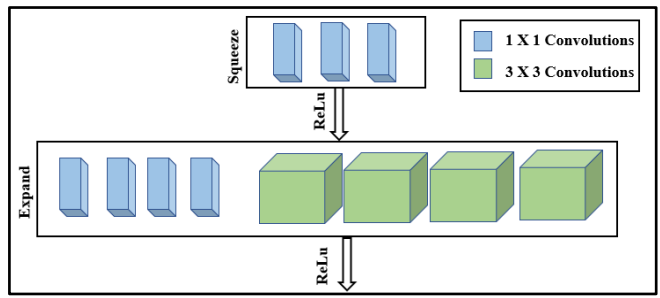

FIGURE 3. SqueezeNet Fire Module with squeeze and expansion Layers

| Replace 3x3 filters with 1x1 filters | • 1x1 filters are used to reduce parameter count while maintaining model performance. 1x1 filter has 9 times fewer parameters compared to 3x3 filters |
|---|---|
| Reduced inputs to 3x3 filters | • Reduced inputs to convolutional layers result in fewer number of parameters. This is achieved by 1X1 filters before 3X3 filters. |
| Delay down sampling in the network | • Down sampling is postponed to later stages such that convolution layers have large activation maps leading to higher classification accuracy. |

FIGURE 4. SqueezeNet Design Strategy

*SqueezeNet1.0 and SqueezeNet1.1*

SqueezeNet1.0 introduced a compact architecture, as shown in Figure 5. It consists of an initial convolutional layer (conv1), eight fire modules (fire2 to fire9), and a final convolutional layer (conv10). The architecture incorporates max-pooling after the initial convolutional layer, after fire modules 4 and 8, and at the end of the network. To improve efficiency, SqueezeNet1.0 employs strategies such as replacing 3x3 filters with 1x1 filters, applying 1x1 filters before 3x3 filters, and delaying down sampling to retain larger activation maps in earlier layers. Building on this, SqueezeNet1.1 (Figure 5) achieves 2.4x less computation [51] and slightly fewer parameters than SqueezeNet1.0 without sacrificing accuracy. It reduces the filter count in conv1 and the squeeze layers while maintaining the same sequence of eight fire modules. Max-pooling is performed after the initial convolutional layer, after fire modules 3 and 5, and at the end of the network. Both architectures demonstrate the effectiveness of parameter-efficient design and are well-suited for deployment on resource-constrained devices.

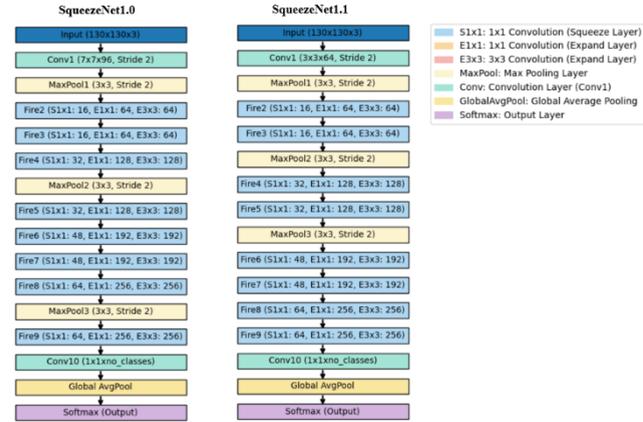

FIGURE 5. SqueezeNet1.0 and SqueezeNet1.1 Architecture

*SqueezeNet Variants Architectures*

To further explore the efficiency of CNN architectures with even fewer parameters, we introduced novel variants of SqueezeNet by reducing the number of fire modules:

**SqueezeNet1.1 Variant 1 (One Fire Module)**
This ultra-lightweight variant, shown in Figure 1, consists of a 3x3 convolutional layer (stride 2, 64 filters) followed by a 3x3 max pooling layer (stride 2). A single fire module is included, which contains a Squeeze Layer (1x1 convolution) and an Expand Layer branching into 1x1 and 3x3 convolutions (both with 64 filters). After the fire module, a 3x3 max pooling layer (stride 2) is applied. The classification layers include a 1x1 convolutional layer, global average pooling, and a softmax output.

**SqueezeNet1.1 Variant 2 (Two Fire Modules)**
This variant, as shown in Figure 1, provides a balance between efficiency and representational power. It begins with a 3x3 convolutional layer (stride 2, 64 filters) followed by a 3x3 max pooling layer (stride 2). Two fire modules are included, each comprising a Squeeze Layer (1x1 convolution) and an Expand Layer with 1x1 and 3x3 branches (both with 64 filters). After the fire modules, a 3x3 max pooling layer (stride 2) is applied. The model concludes with a 1x1 convolutional layer, global average pooling, and a softmax activation. This configuration offers improved feature extraction compared to Varian 1 while remaining computationally efficient.

**SqueezeNet1.1 Variant 3 (Four Fire Modules)**
This variant, as shown in Figure 1, enhances feature representation while keeping the architecture lightweight. It begins with a 3x3 convolutional layer (stride 2, 64 filters) followed by a 3x3 max pooling layer (stride 2). Four fire modules are included. Each fire module consists of a Squeeze Layer (1x1 convolution) and an Expand Layer with two branches: a 1x1 convolution and a 3x3 convolution. The first two fire modules have 64 filters in their branches, while the next two fire modules have 128 filters. A 3x3 max pooling layer (stride 2) is applied after the fire Modules, followed by a 1x1 convolutional layer, global average pooling, and a softmax activation. This variant can be ideal for applications requiring higher accuracy without a significant increase in computational cost.

**Training Parameters**
The details of each architecture's trainable parameters and size (8-bitwise) are shown in Figure 6. Our proposed variants, including this one, aim to optimize computational efficiency and accuracy by selectively reducing the number of fire modules while retaining sufficient representational capacity. The Variant 1, with 13,458 parameters (52.57 KB/0.051MB), achieves a 54x times reduction in training parameters compared to the original SqueezeNet1.1 (723,522 parameters, 2.76 MB). Similarly, Variant 2 has 25,890 parameters (101.13 KB/0.099MB), offering a 28x reduction, while Variant 3, with 120,930 parameters (472.38 KB/0.461MB), is 6x smaller than the original model. We have comprehensively evaluated these architectures to identify the trade-offs between accuracy and computational resource requirements. These variants are particularly suited for highly resource-constrained environments, such as mobile devices, edge computing platforms, and other scenarios where processing power and memory are limited.

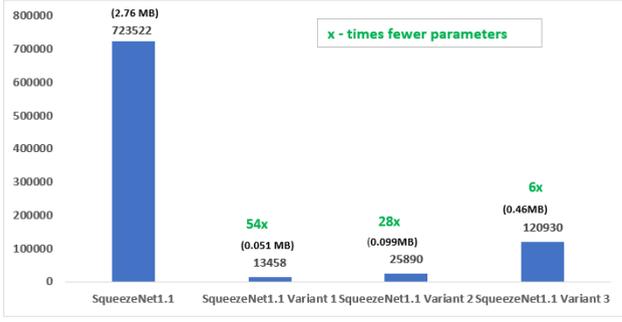

**FIGURE 6.** Training parameters for SqueezeNet1.1 and variant architecture

**Evaluation**

We conducted a performance evaluation of our proposed variant models in comparison to the original SqueezeNet1.1, focusing on their effectiveness in malaria detection. The assessment utilized key metrics such as accuracy, precision, recall, F1 score, area under the ROC curve, confusion matrix, model size, and training and inference times. These metrics offered a comprehensive analysis of each model's ability to differentiate between malaria-positive and normal cases, providing insights into class-specific performance, overall classification accuracy, and computational efficiency. This detailed evaluation enabled a clear comparison between SqueezeNet1.1 and its optimized variants.

**Accuracy**

Accuracy is a critical metric that evaluates the proportion of correct predictions, encompassing both true positives (TP) and true negatives (TN), relative to the total predictions made. It serves as an overall indicator of a model's performance, as defined in Equation 1. The components used to calculate accuracy include True Positives (TP), True Negatives (TN), False Positives (FP), and False Negatives (FN).

$$Accuracy = \frac{TP+TN}{TP+TN+FP+FN} \quad (1)$$

**Confusion Matrix**

The confusion matrix provides a comprehensive summary of the model's performance for each class, detailing the counts of True Positives, True Negatives, False Positives, and False Negatives. It is an essential metric for analyzing class-specific performance and detecting issues like class imbalances or misclassifications.

**Precision, Recall, and F1 Score**

Precision, Recall, and F1 Score provide valuable class-specific insights, particularly for assessing model performance on imbalanced datasets. These metrics are derived from the confusion matrix. Precision (Equation 2) represents the ratio of correctly predicted positive cases (True Positives) to all predicted positives, indicating the reliability of positive predictions. Recall (Equation 3) measures the ratio of correctly predicted positives to all actual positives, highlighting the model's ability to detect malaria cases. The F1 Score (Equation 4), which is the harmonic mean of Precision and Recall, offers a comprehensive measure of the model's effectiveness in detecting malaria.

$$Precision = \frac{TP}{TP+FP} \quad (2)$$

$$Recall = \frac{TP}{TP+FN} \quad (3)$$

$$F1 = 2 X \frac{Precision \; X \; Recall}{Precision+Recall} \quad (4)$$

Weighted Average

In classification metrics, weighted averages are often used to aggregate Precision, Recall, and F1 Scores across multiple classes, especially in imbalanced datasets. This approach provides a more comprehensive assessment of model performance by factoring in both class imbalance and the performance of individual classes, offering insights beyond single-class metrics. The weighted average (Equation 5) is the mean of the metrics (Precision, Recall, F1 Score) for each class, weighted by the support or the number of instances in each class. This method accounts for class distribution, delivering a more accurate representation of model performance in imbalanced datasets. Larger classes have a greater impact on the weighted average, making it an effective metric for overall model evaluation. To calculate the weighted average, each metric $M_i$ for class $i$ is multiplied by the support $n_i$, the number of instances in the $i$-th class, and the sum of these weighted values is divided by the total number of instances $N$ in the dataset.

$$Weighted \; Average = \frac{1}{N} \sum_{i=1}^{c} M_i \times n_i \quad (5)$$

**Area Under the Curve (AUC) and Receiver Operating Characteristic (ROC) Curve**

The AUC score quantifies the model's ability to differentiate between malaria and normal cases. It is derived from the ROC curve, which plots the True Positive Rate (TPR) against the False Positive Rate (FPR) for various classification thresholds. An AUC score of 1.0 indicates perfect classification, while a score of 0.5 suggests no better than random guessing. The AUC is calculated as shown in Equation 6. The ROC curve illustrates the diagnostic performance of a binary classifier by plotting the TPR versus the FPR as the classification threshold varies. The TPR and FPR are calculated using Equations 7 and 8.

$$AUC = \int_0^1 TPR(t) \; d(FPR(t)) \quad (6)$$

$$TPR = \frac{TP}{TP+FN} \quad (7)$$

$$FPR = \frac{FP}{FP+TN} \quad (8)$$

## IV. RESULTS

We performed the experiments on Google Colab, utilizing CPU resources, 51 GB of RAM, and 225.8 GB of disk space. To implement our proposed Squeezenet variant models, we utilized Python 3, along with essential libraries such as Scikit-Learn, Matplotlib, Keras, and TensorFlow. We have taken the original Squeezenet1.1 code from authors Forrest Iandola and Christopher Masch's Github repositories [52][53] and further modified the code for proposed variants based on the details included in the methodology section. We have made our Squeezenet1.1 Variant's code available in the GitHub repository [54]. We trained Squeezenet1.1 and variants with a malaria training dataset of 22046 images.

For model compilation, we used the Adam optimizer with a learning rate of 1e-4, paired with a categorical cross-entropy loss function. Training was carried out on augmented data for 100 epochs with a batch size of 32. After the training we compared the performance of the SqueezeNet1.1 variants and the original SqueezeNet1.1 model, using a validation dataset of 5512 images, consisting of 2756 malaria infected and 2756 non-infected images. The models were evaluated based on various metrics outlined in the methodology section.

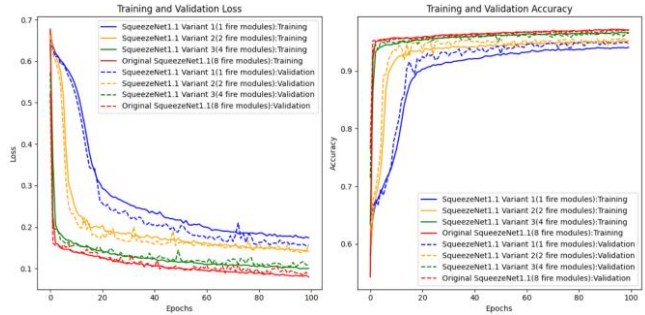

FIGURE 7. Training, Validation Loss and Accuracy

The original SqueezeNet1.1 demonstrates the most efficient convergence, as shown in Figure 7, achieving the lowest loss values for both training and validation sets. Variant 3 closely follows, showing a smooth and stable decline in loss, nearly matching the performance of the SqueezeNet1.1. Variants 1 and 2, however, exhibit slightly higher loss values, with Variant 1 converging more slowly and displaying the highest overall validation loss among the models. The SqueezeNet1.1 achieves the highest accuracy, stabilizing at a near-perfect level. Variant 3 again displays competitive performance, with validation accuracy closely mirroring that of the SqueezeNet1.1. Variants 1 and 2 show moderate performance, with Variant 1 lagging behind the most. Notably, all models exhibit well-aligned training and validation curves, indicating minimal overfitting and strong generalization capabilities across the board. In our study, we observed a clear trend toward model convergence across all SqueezeNet1.1 and variants, as indicated by the stabilization of both training and validation loss over the course of training. The validation accuracy either closely matches or slightly exceeds the training accuracy in most cases, underscoring the model's ability to maintain strong predictive performance while being well-regularized and generalizing well to unseen data.

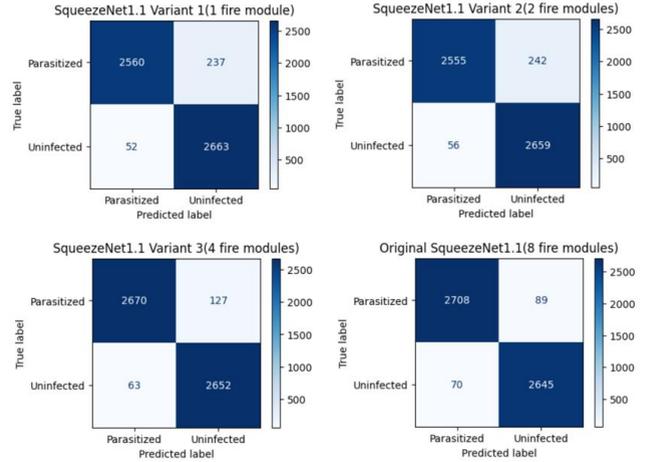

FIGURE 8. Confusion Matric for SqueezeNet1.1 and its variants

The confusion matrices in Figure 8 illustrate the classification performance of the original SqueezeNet1.1 and its three variants with reduced fire modules. We observe from the confusion matrix that Variant 1 correctly identifies 2,560 parasitized samples and 2,663 uninfected samples while misclassifying 237 parasitized samples as uninfected and 52 uninfected samples as parasitized. Variant 2 accurately classifies 2,555 parasitized samples and 2,659 uninfected samples. However, it mislabeled 242 parasitized samples as uninfected and 56 uninfected samples as parasitized. Variant 3 identifies 2,670 parasitized samples and 2,652 uninfected samples, misclassifying 127 parasitized samples and 63 uninfected samples. The reduced number of misclassifications demonstrates enhanced performance with the slightly added complexity. The SqueezeNet1.1 exhibits the highest accuracy among the variants. It correctly classifies 2,708 parasitized samples and 2,645 uninfected samples while only misclassifying 89 parasitized samples and 70 uninfected samples.

Among all models, the SqueezeNet1.1 demonstrates superior performance with the highest true positive counts for both Parasitized and Uninfected classes, reflecting its robustness in classifying malaria-infected cells. Variant 3 achieves results comparable to the SqueezeNet1.1, with slightly fewer misclassifications in the Parasitized class while maintaining high accuracy for the Uninfected class. Variants 1 and 2 exhibit higher misclassification rates, particularly for the Parasitized class, compared to the SqueezeNet1.1 and Variant 3.

| Model | Class | Precision | Recall | F1-score |
|---|---|---|---|---|
| SqueezeNet1.1 Variant 1 (1 fire module) | Parasitized | 98.01% | 91.53% | 94.66% |
| | Uninfected | 91.83% | 98.08% | 94.85% |
| SqueezeNet1.1 Variant 2 (2 fire module) | Parasitized | 97.86% | 91.35% | 94.49% |
| | Uninfected | 91.66% | 97.94% | 94.69% |
| SqueezeNet1.1 Variant 3 (4 fire module) | Parasitized | 97.69% | 95.46% | 96.56% |
| | Uninfected | 95.43% | 97.68% | 96.54% |
| Original SqueezeNet1.1 (8 fire module) | Parasitized | 97.48% | 96.82% | 97.15% |
| | Uninfected | 96.74% | 97.42% | 97.08% |

**TABLE 1: Brief Class level Precision, Recall, and F1-Score**

The precision, recall, and F1-scores at the class level for the original SqueezeNet1.1 and its variants offer detailed insights into their classification performance, as shown in Table 1. The SqueezeNet1.1 stands out with the best overall performance, achieving F1-scores of 0.97 for both the Parasitized and Uninfected classes. This model excels at balancing precision and recall, especially for the Parasitized class, where it maintains the highest recall (96.82%) while demonstrating solid precision (97.48%). Variant 3 performs nearly identically to the SqueezeNet1.1 with an F1-score of 0.96 for both classes. Variant 2 shows a slight decline in performance compared to Variant 3 but still outperforms Variant 1. It achieves an F1-score of 0.94 for both classes, with a strong precision of 97.86% for the Parasitized class and 91.66% for the Uninfected class. Variant 1, though the simplest and smallest model, demonstrates reasonable performance with an F1-score of 0.94 for the Uninfected class but a lower F1-score of 0.92 for the Parasitized class. This lower F1-score reflects a decrease in recall, indicating that while the model offers strong precision, it is less effective at identifying all instances of the Parasitized class.

| Model | Accuracy | Precision (weighted avg) | Recall (weighted avg) | F1-score (weighted avg) |
|---|---|---|---|---|
| SqueezeNet1.1 Variant 1 (1 fire module) | 94.76% | 94.96% | 94.76% | 94.75% |
| SqueezeNet1.1 Variant 2 (2 fire modules) | 94.59% | 94.80% | 94.59% | 94.59% |
| SqueezeNet1.1 Variant 3 (4 fire modules) | 96.55% | 96.58% | 96.55% | 96.55% |
| Original SqueezeNet1.1 (8 fire modules) | 97.12% | 97.12% | 97.12% | 97.12% |

**TABLE 2: Comparison of SqueezeNet1.1 and variants performance**

The accuracy, precision, recall, and F1-scores (weighted averages) for the variants and the original SqueezeNet1.1 offer valuable insights into their overall classification capabilities, as shown in Table 2. The SqueezeNet1.1 delivers the best overall performance, achieving an accuracy of 97.12% along with a precision, recall, and F1-score of 97.12%. Variant 3 follows closely with an accuracy of 96.55% and equally impressive weighted average scores of 96.58% for precision, 96.55% for recall, and 96.55% for F1-score. Variant 2 shows a slight decline compared to Variant 3, with an accuracy of 94.59%, precision of 94.80%, recall of 94.59%, and an F1-score of 94.59%. Variant 1, while maintaining high performance, is the least effective among the variants, with an accuracy of 94.76%, precision of 94.96%, recall of 94.76%, and F1-score of 94.75%.

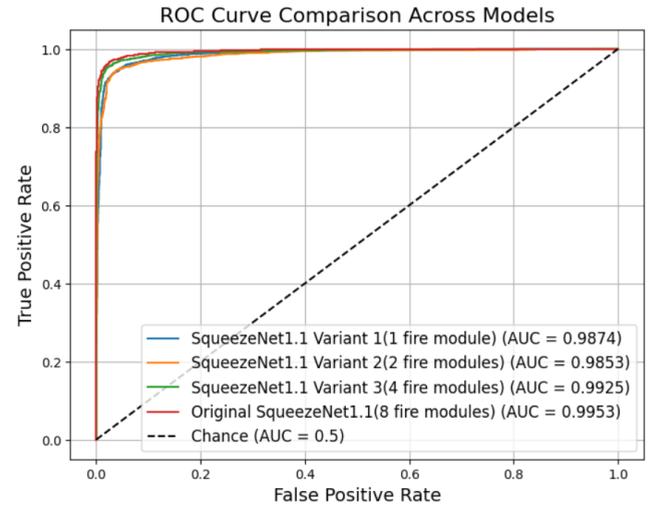

**FIGURE 9. ROC curve comparison for SqueezeNet1.1 and its variants**

The ROC curve analysis shown in Figure 9 highlights the strong classification performance of the original SqueezeNet1.1 and its variants. The AUC values are consistently high, with SqueezeNet1.1 achieving the best performance at an AUC of 0.9953. Among the variants, Variant 3 follows closely with an AUC of 0.9925, while Variants 1 and 2 achieve slightly lower AUC values of 0.9874 and 0.9853, respectively. These results demonstrate that all models excel in distinguishing between the two classes, as evidenced by their proximity to the ideal top-left corner of the ROC space. The higher AUC values for SqueezeNet1.1 and Variant 3 suggest superior discriminative capabilities. The close clustering of curves further indicates consistent performance across the models, with minimal compromise in classification quality, even in simpler variants.

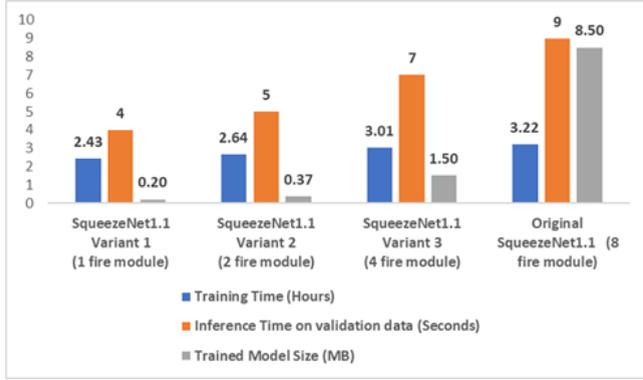

FIGURE 10. Training, Validation Time and Model size of SqueezeNet1.1 and Its variants

The comparative analysis of training time, inference time on validation data (5512 images) and model size shown in Figure 10 reveals significant differences between the original SqueezeNet1.1 and its variants. The SqueezeNet1.1 exhibits the highest resource requirements, with a training time of 3.22 hours, an inference time of 9 seconds, and a model size of 8.5 MB. In contrast, the simplified variants demonstrate considerable reductions in all metrics. For instance, Variant 1 reduces the training time by 24.5% (2.43 hours), inference time by 55.6% (4 seconds), and model size by 97.6% (0.2 MB) compared to SqueezeNet1.1. Similarly, Variant 2 achieves a 17.7% (2.64 hours) reduction in training time, a 44.4% (5 seconds) decrease in inference time, and a 95.6% (0.37MB) reduction in model size to that of SqueezeNet1.1. Variant 3 strikes a balance between efficiency and performance, reducing the training time by 6.5% (3 hours), inference time by 22.2% (7 seconds), and model size by 82.4% (1.5 MB) to that of SqueezeNet1.1 while maintaining competitive predictive capabilities. These results highlight the potential for deploying compact models like Variants 1 and 2 in resource-constrained environments without substantial performance degradation. Also note that while the model size (storage in MB) and training parameter count both represent aspects of the model, they measure different attributes. Model size reflects the storage space needed, while parameter count correlates with the computational load during training and inference.

## V. Discussions

Overall, the results of our study underscore the potential of SqueezeNet1.1variants as effective architectures for malaria detection. From our comprehensive evaluation, the original SqueezeNet1.1 achieves slightly better results compared to Variant 3. However, this advantage is accompanied by increased computational demands, as indicated by longer training and inference times and a larger model size. Among the variants, Variant 3 proves to be a competitive alternative, demonstrating an effective balance between classification performance and computational efficiency. Variant 3 results closely match SqueezeNet1.1 in accuracy and other evaluation metrics. Variant 3's reduced model complexity and faster training and inference times make it particularly well-suited for deployment in resource-constrained environments. This highlights the potential of judicious architectural simplifications to achieve both efficiency and strong predictive capability. Variants 1 and 2 showcase an explicit trade-off between performance and computational demands. These models, characterized by further reduced model size and complexity, demonstrate promising results but exhibit slightly lower accuracy and higher misclassification rates compared to Variant 3 and SqueezeNet1.1. Nevertheless, their lightweight design and significantly faster processing times make them attractive for very high resource-constrained environments. Despite their simplicity, both variants maintain satisfactory performance, further validating the adaptability of SqueezeNet-based architectures to meet varying computational resource limitations.

Our analysis further reveals that while increasing the number of fire modules in the architecture enhances classification performance, the benefits plateau beyond a certain point. This observation suggests diminishing returns with additional complexity, emphasizing the importance of balancing performance gains against computational overhead. Variant 3 exemplifies this balance by achieving competitive accuracy with fewer fire modules, thereby offering a more efficient solution without significant compromise in malaria classification performance.

In comparison to other deep learning architectures, such as ResNet-50 and VGG-16, our proposed ultra-lightweight variants demonstrate remarkable computational efficiency. ResNet-50 and VGG-16, though powerful, require substantially larger model sizes and longer training times, making them less practical for deployment in low-resource environments. In contrast, SqueezeNet1.1, a streamlined version of SqueezeNet1.0, reduces computational demands by 2.4x. The further optimizations in our proposed variants build on this efficiency, providing even more compact and resource-friendly solutions. Overall, the study highlights the adaptability of SqueezeNet1.1 and its variants for malaria detection, offering a spectrum of solutions that cater to diverse resource constraints. These findings emphasize the importance of tailoring model complexity to application-specific requirements, ensuring an optimal balance between computational efficiency and predictive performance.

## VI. Limitations

Despite the promising results, our study has limitations that need to be addressed in future work. The model was trained and evaluated on a single dataset, the Kaggle Malaria dataset (sourced from NIH). While this dataset is comprehensive and large, it may not fully capture the diverse image conditions and parasite variations encountered in different geographic regions. Variations in microscope equipment, staining protocols, and local parasite species could impact the model's performance when applied to new data sources. Additionally, the dataset comprises images with a balanced distribution of parasitized and uninfected cells, which may not reflect the actual prevalence rates in clinical settings, where uninfected cells are often more prevalent. Incorporating real-time testing with medical professionals could provide critical insights into the models' practical utility and usability, which should be considered in future work.

Although the models were trained and evaluated on a single malaria dataset that meets our research objectives, there is significant potential to explore our approach with other medical imaging tasks or datasets to assess its generalizability. Future work should investigate the performance of SqueezeNet variants on diverse datasets to evaluate their broader applicability. Additionally, alternative image processing techniques and model training hyperparameters can be explored in future work to further enhance performance and adaptability to various medical imaging challenges. Our study did not assess the models in real-time diagnostic settings. Furthermore, our study focused exclusively on binary classification (parasitized vs. non-parasitized). Extending the models to multi-class classification tasks, such as differentiating between different diseases or stages of infection, could improve their clinical utility. Like many deep learning models, the SqueezeNet variants function as black-box models, offering limited interpretability of their decision-making processes. Future research could incorporate explainable AI techniques to enhance understanding and trust in the model's predictions

## VII. CONCLUSION

Our study demonstrates the computational effectiveness of our novel ultralight weight SqueezeNet1.1 architecture variants for malaria detection. The results confirm that the original SqueezeNet1.1 model, with its 8 fire modules, offers the best classification performance but at the cost of increased computational demands. On the other hand, Variant 3 (with four fire modules) achieves nearly identical accuracy with a 6x reduction in computational overhead and significant reduction in training, inference time, making it a strong candidate for deployment in environments with limited computational resources. Variants 1(one fire module) and 2 (two fire modules), despite their slightly lower classification accuracy, offer substantial reductions in computational overhead, with Variant 2 reducing computational demands by 28x and Variant 1 achieving a 54x reduction compared to SqueezeNet1.1. Variants 1 and 2 offer substantial reductions in training and inference time, making them suitable for scenarios with extreme resource constraints.

While these models show promising results in a binary classification task, future research could extend them to multi-class classification or other medical imaging tasks. Additionally, improving the interpretability of these models through explainable AI techniques could enhance their clinical applicability. Overall, our study highlights the potential of our SqueezeNet1.1 architecture variants for malaria detection, offering a flexible approach that allows for the selection of the most suitable model based on resource constraints, balancing complexity and performance.

**SURESH BABU NETTUR** has received his Master of Science (M.S) degree from the Birla Institute of Technology and Science (BITS), Pilani, Rajasthan, India, and his Bachelor of Technology degree in Computer Science and Engineering from Nagarjuna University, Guntur, India. With over two decades of expertise, he has established himself as a thought leader in artificial intelligence, deep learning, and machine learning solutions. His work spans the development of scalable and intelligent systems across industries such as healthcare, finance, telecom, and manufacturing. Suresh has been at the forefront of integrating AI-driven innovations into real-world applications, leveraging cutting-edge technologies such as OpenAI models, GitHub Copilot, and custom deep learning architectures to deliver transformative solutions. His contributions include designing and implementing advanced machine learning models, optimizing deep learning architectures for resource-constrained environments, and integrating AI solutions into software development and testing pipelines. With significant experience in cross-functional team leadership and managing onsite-offshore collaboration models, he has successfully delivered AI-powered applications across cloud platforms like AWS.

Suresh is passionate about applying AI to solve complex problems in healthcare and finance, including predictive analytics, automation, and intelligent decision-making systems. He is proficient in Agile methodologies, Test-Driven Development (TDD), and service-oriented architectures (SOA), ensuring seamless integration of AI and machine learning into software systems. As an advocate for innovative AI applications, Suresh is committed to advancing the field through sustainable and impactful solutions that redefine industry standards and improve quality of life.

**SHANTHI KARPURAPU** received the Bachelor of Technology degree in chemical engineering from Osmania University, Hyderabad, India and the Masters technology degree in chemical engineering from Institute of Chemical Technology, Mumbai, India.
She has over a decade of experience leading, designing, and developing test automation solutions for various platforms across healthcare, banking, and manufacturing industries using Agile and Waterfall methodologies. She is experienced in building reusable and extendable automation frameworks for web applications, REST, SOAP, and microservices. She is a strong follower of the shift-left testing approach, a certified AWS Cloud practitioner, and a machine learning specialist. She is passionate about utilizing AI-related technologies in software testing and the healthcare industry.

**UNNATI NETTUR** currently pursuing an undergraduate degree in Computer Science at Virginia Tech, Blacksburg, VA, USA. She possesses an avid curiosity about the constantly evolving field of technology and software development, with a particular interest in Artificial Intelligence. She is passionate about gaining experience in building innovative and creative solutions for current issues in the field of software engineering.

**LIKHIT SAGAR GAJJA** pursuing a Computer Science Bachelor's degree at BML Munjal University, Haryana, INDIA. He is evident in showing his passion for the dynamic field of technology and software development. His specific interests include Artificial Intelligence, Prompt Engineering, and Game Designing technologies, highlighting his dedication to obtaining hands-on experience and developing innovative solutions for real-time issues in software engineering.

**SRAVANTHY MYNENI** earned master of science in information technology and management from Illinois institute of technology, Chicago, Illinois in 2017 and Bachelor's degree in computer science in 2013. She is currently working as an engineer focused on data engineering and analysis. She has 8+ years of experience in designing, building and deploying data centric solutions using Agile and Waterfall methodologies. She is enthusiastic about data analysis, data engineering and AI application to provide solutions for real world problems.

**AKHIL DUSI** currently pursuing Masters of Information Sciences at University of Indiana Tech, Indiana, USA. He is a passionate researcher and developer with a diverse background in software development, cybersecurity, and emerging technologies. He has a proven ability to deliver innovative and practical solutions. His work includes developing web and mobile-based applications, conducting vulnerability assessments and penetration testing, and leveraging cloud platforms for efficient infrastructure management. He is certified in cybersecurity and machine learning, reflecting a strong commitment to continuous learning and staying at the forefront of technological advancements. His research interests focus on artificial intelligence, IoT, and secure system design, with a vision to drive impactful innovations.

**LALITHYA POSHAM** is an MBBS graduate from Nanjing Medical University. She has a strong passion for advancing clinical research and improving patient outcomes. With a solid foundation in medical education, she is particularly interested in exploring innovative diagnostic approaches and aims to integrate clinical expertise with research to drive improvements in healthcare systems.